\documentclass[11pt,a4paper]{article}

\usepackage{colacl}

\usepackage{linguex}
\usepackage{epsfig}
\usepackage{hyperref}

\title{A Model for Fine-Grained Alignment of Multilingual Texts}

\author{%
  Lea CYRUS and Hendrik FEDDES\thanks{We would like to thank our colleague
    Frank Schumacher for many valuable comments on this paper.}\\
  Arbeitsbereich Linguistik\\
  University of M\"{u}nster\\
  H\"{u}fferstra{\ss}e 27, 48149 M\"{u}nster, Germany\\
  \{lea,feddes\}@marley.uni-muenster.de%
}

\begin{document}


\newcommand{\fuse}[0]{FuSe}
\newcommand{\dbliteral}[1]{\texttt{#1}}


\maketitle

\begin{abstract}
  While alignment of texts on the sentential level is often seen as being too
  coarse, and word alignment as being too fine-grained, bi- or multilingual
  texts which are aligned on a level in-between are a useful resource for many
  purposes.  Starting from a number of examples of non-literal translations,
  which tend to make alignment difficult, we describe an alignment model which
  copes with these cases by explicitly coding them. The model is based on
  predicate-argument structures and thus covers the middle ground between
  sentence and word alignment.  The model is currently used in a recently
  initiated project of a parallel English-German treebank (\fuse), which can
  in principle be extended with additional languages.
\end{abstract}

\section{Introduction}
\label{sec:intro}

When building parallel linguistic resources, one of the most obvious problems
that need be solved is that of alignment. Usually, in sentence- or
word-aligned corpora, alignments are unmarked relations between corresponding
elements. They are unmarked because the kind of correspondence between two
elements is either obvious or beyond classification. E.\,g., in a
sentence-aligned corpus, the $n:m$ relations that hold between sentences
express the fact that the propositions contained in $n$ sentences in L1 are
basically the same as the propositions in $m$ sentences in L2 (lowest common
denominator). No further information about the kind of correspondence could
possibly be added on this degree of granularity. On the other hand, in
word-aligned corpora, words are usually aligned as being ``lexically
equivalent'' or are not aligned at all.\footnote{Cf.  the approach described
  in \cite{Melamed98}.}  Although there are many shades of ``lexical
equivalence'', these are usually not explicitly categorised. As
\cite{HansenNeumann03} point out, for many research questions neither type of
alignment is sufficient, since the most interesting phenomena can be found on
a level between these two extremes.

We propose a more finely grained model of alignment which is based on
monolingual predicate-argument structures, since we assume that, while
translations can be non-literal in a variety of ways, they must be based on
similar predicates and arguments for some kind of translational equivalence to
be achieved.  Furthermore, our model explicitly encodes the ways in which the
two versions of a text deviate from each other. \cite{Salkie02} points out
that the possibility to investigate what types of non-literal translations
occur on a regular basis is one of the major profits that linguists and
translation theorists can draw from parallel corpora.

In Section~\ref{sec:differences}, we begin by describing some ways in which
translations can deviate from one another. We then describe in detail the
alignment model, which is based on a monolingual predicate-argument structure
(Section~\ref{sec:model}). In Section~\ref{sec:outlook} we conclude by
introducing the parallel treebank project \fuse{} which uses the model
described in this paper to align German and English texts from the Europarl
parallel corpus \cite{Koehn02}.

\section{Differences in Translations}
\label{sec:differences}

In most cases, translations are not absolutely literal counterparts of their
source texts. In order to avoid translationese, i.\,e. deviations from the
norms of the target language, a skilled translator will apply certain
mechanisms, which \cite{Salkie02} calls ``inventive translations'' and which
need to be captured and systematised. The following section will give some
examples\footnote{As we work with English and German, all examples are taken
  from these two languages. They are taken from the Europarl corpus (see
  Section~\ref{sec:outlook}) and are abbreviated where necessary.
  Unfortunately, it is not easily discernible from the corpus data which
  language is the source language.  Consequently, our use of the terms
  'source', 'target', 'L1', and 'L2' does not admit of any conclusions as to
  whether one of the languages is the source language, and if so, which one.}
of common discrepancies encountered between a source text and its translation.

\subsection{Nominalisations}
\label{sec:nominalisations}

Quite frequently, verbal expressions in L1 are expressed by corresponding
nominalisations in L2. This departure from the source text results in a
completely different structure of the target sentence, as can be seen in
\ref{ex:harmonise} and \ref{ex:harmonisierung},  where the English verb
\emph{harmonise} is expressed as \emph{Harmonisierung} in German. The argument
of the English verb functioning as the grammatical subject is realised as a
postnominal modifier in the German sentence.

\ex. The laws against racism must be
harmonised.\footnote{Europarl:de-en/ep-00-01-19.al, 489.}
\label{ex:harmonise}

\exg.  Die Harmonisierung der Rechtsvorschriften gegen den Rassismus ist
dringend erforderlich.\\
The harmonisation of\_the laws against the racism is urgently necessary.\\
\label{ex:harmonisierung}

This case is particularly interesting, because it involves a case of modality.
In the English sentence, the verb is modified by the modal auxiliary
\emph{must}. In order to express the modality in the German version, a
different strategy is applied, namely the use of an adjective with modal
meaning (\emph{erforderlich}, 'necessary'). Consequently, there are two
predications in the German sentence as opposed to only one predication in the
English sentence.

\subsection{Voice}
\label{sec:voice}

A further way in which translations can differ from their source is the choice
of active or passive voice. This is exemplified by \ref{ex:passive-e} and
\ref{ex:passive-g}. Here, the direct object of the English sentence
corresponds to the grammatical subject of the German sentence, while the
subject of the English sentence is realised as a prepositional phrase with
\emph{durch} in the German version.

\ex. The conclusions of the Theato report safeguard them
perfectly.\footnote{Europarl:de-en/ep-00-01-18.al, 749.}
\label{ex:passive-e}

\exg. Durch die Schlu{\ss}folgerungen des Berichts Theato werden sie
uneingeschr\"{a}nkt bewahrt.\\
By the conclusions of\_the report Theato are they unlimitedly safeguarded\\
\label{ex:passive-g}

\subsection{Negation}
\label{sec:negation}

Sometimes, a positive predicate expression is translated by negating its
antonym. This is the case in \ref{ex:inapplicable} and \ref{ex:anwendbar}:
both sentences contain a negative statement, but while the negation is
incorporated into the English adjective by means of the negative prefix
\emph{in-}, it is achieved syntactically in the German sentence.

\ex. the Directive is inapplicable in
Denmark\footnote{Europarl:de-en/ep-00-01-18.al, 2522.}
\label{ex:inapplicable}

\exg. die Richtlinie ist in D\"{a}nemark nicht anwendbar\\
the Directive is in Denmark not applicable\\
\label{ex:anwendbar}

\subsection{Information Structure}
\label{sec:information}

Sentences and their translations can be organised differently with regard to
their information structure. Sentences \ref{ex:motion} and \ref{ex:mitgeben}
are a good example for this type of non-literal translation.

\ex. Our motion will give you a great deal of food for thought,
Commissioner\footnote{Europarl:de-en/ep-00-01-18.al, 53.}
\label{ex:motion}

\exg. Eine Reihe von Anregungen werden wir Ihnen, Herr Kommissar, mit unserer
Entschlie{\ss}ung mitgeben\\
A row of suggestions will we you, Mr. Commissioner, with our resolution give\\
\label{ex:mitgeben}

The German sentence is rather inconspicuous, with the grammatical subject
being a prototypical agent (\emph{wir}, 'we'). In the English version,
however, it is the means that is realised in subject position and thus
perspectivised. The corresponding constituent in German (\emph{mit unserer
  Entschlie{\ss}ung}, 'with our motion') is but an adverbial. In English, the
actual agent is not realised as such and can only be identified by a process
of inference based on the presence of the possessive pronoun \emph{our}. Thus,
while being more or less equivalent in meaning, this sentence pair differs
significantly in its overall organisation.

\section{Alignment Model}
\label{sec:model}

The alignment model we propose is based on the assumption that a
representation of translational equivalence can best be approximated by
aligning the elements of monolingual predicate-argument structures.
Section~\ref{sec:pa} describes this layer of the model in detail and shows how
some of the differences in translations described in
Section~\ref{sec:differences} can be accomodated on such a level.  We assume
that the annotation model described here is an extension to linguistic data
which are already annotated with phrase-structure trees, i.\,e. treebanks.
Section~\ref{sec:binding} shows how the binding of predicates and arguments to
syntactic nodes is modelled.  Section~\ref{sec:align} describes the details of
the alignment layer and the tags used to mark particular kinds of alignments,
thus accounting for some more of the differences shown in
Section~\ref{sec:differences}.

\subsection{Predicates and Arguments}
\label{sec:pa}

The predicate-argument structures used in our model consist solely of
predicates and their arguments. Although there is usually more than one
predicate in a sentence, no attempt is made to nest structures or to join the
predications logically in any way. The idea is to make the predicate-argument
structure as rich as is ne\-cessary to be able to align a sentence pair while
keeping it as simple as possible so as not to make it too difficult to
annotate.  In the same vein, quantification, negation, and other operators are
not annotated. In short, the predicate-argument structures are not supposed to
capture the semantics of a sentence exhaustively in an interlingua-like
fashion.

To have clear-cut criteria for annotators to determine what a predicate is, we
rely on the heuristic assumption that predicates are more likely to be
expressed by tokens belonging to some word classes than by tokens belonging to
others.  Potential predicate expressions in this model are verbs, deverbal
adjectives and nouns\footnote{For all non-verbal predicate expressions for
  which a derivationally related verbal expression exists it is assumed that
  they are deverbal derivations, etymological counter-evidence
  notwithstanding.} or other adjectives and nouns which show a syntactic
subcategorisation pattern.  The predicates are represented by the capitalised
citation form of the lexical item (e.\,g.\ \textsc{harmonise}).  They are
assigned a class based on their syntactic form (\emph{v}, \emph{n}, \emph{a}
for 'verbal', 'nominal', and 'adjectival', respectively), and derivationally
related predicates form a predicate group.

Arguments are given short intuitive role names (e.\,g.\ 
\textsc{ent\_harmonised}, i.\,e.\ the entity being harmonised) in order to
facilitate the annotation process.  These role names have to be used
consistently only within a predicate group.  If, for example, an argument of
the predicate \textsc{harmonise} has been assigned the role
\textsc{ent\_harmonised} and the annotator encounters a comparable role as
argument to the predicate \textsc{harmonisation}, the same role name for this
argument has to be used.\footnote{Keeping the argument names consistent for
  all predicates within a group while differentiating the predicates on the
  basis of syntactic form are complementary principles, both of which are
  supposed to facilitate querying the corpus.  The consistency of argument
  names within a group, for example, enables the researcher to analyse
  paradigmatically all realisations of an argument irrespective of the
  syntactic form of the predicate. At the same time, the differentiation of
  predicates makes possible a syntagmatic analysis of the differences of
  argument structures depending on the syntactic form of the predicate.}

\begin{figure*}[p]
  \begin{center}
    \epsfig{file=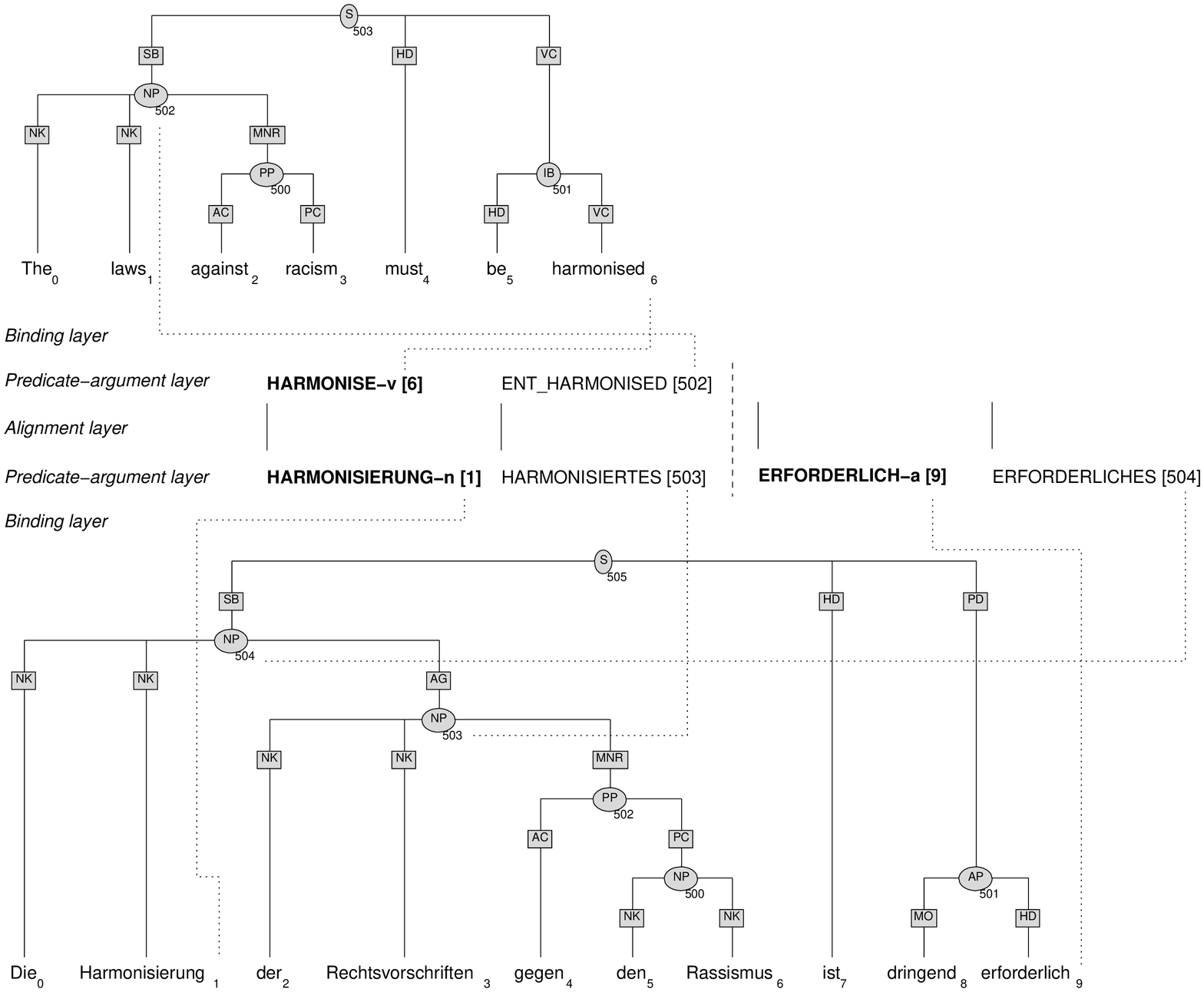,height=.42\textheight}
    \caption{Alignment of a verb/direct-object construction 
      with a noun/modifier construction}
    \label{fig:pa}
  \end{center}
\end{figure*}

The usefulness of such a structure can be shown by analysing the sentence pair
\ref{ex:harmonise} and \ref{ex:harmonisierung} in
Section~\ref{sec:nominalisations}. While the syntactic constructions differ
considerably, the predicate-argument structure shows the correspondence quite
clearly (see the annotated sentences in Figure~\ref{fig:pa}\footnote{All
  figures are at the end of the paper.}): in the English sentence, we find the
predicate \textsc{harmonise} with its argument \textsc{ent\_harmonised}, which
corresponds to the predicate \textsc{harmonisierung} and its argument
\textsc{harmonisiertes} in the German sentence. The information that a
predicate of the class \emph{v} is aligned with a predicate of the class
\emph{n} can be used to query the corpus for this type of non-literal
translations.

\begin{figure*}[p]
  \begin{center}
    \epsfig{file=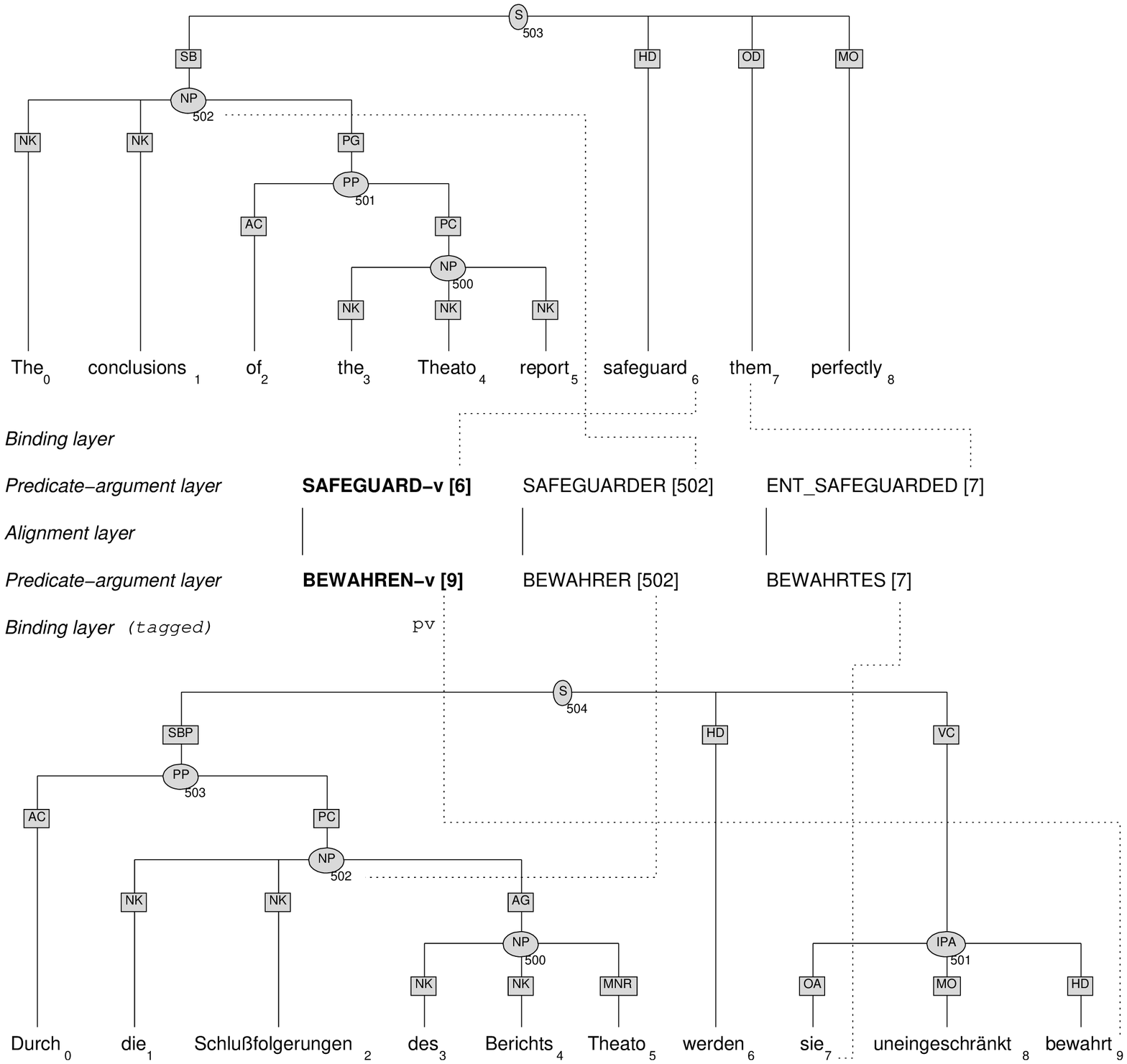,height=.45\textheight}
    \caption{Active vs.\ passive voice in translations: an example of a
      tagged binding (\dbliteral{pv})}
    \label{fig:passive}
  \end{center}
\end{figure*}

The active vs.\ passive translation in sentences \ref{ex:passive-e} and
\ref{ex:passive-g} is another phenomenon which is accomodated by a
predicate-argument structure (Figure~\ref{fig:passive}): the subject
\textsc{np}$_{502}$ in the English sentence corresponds to the passivised
subject \textsc{np}$_{502}$ (embedded in \textsc{pp}$_{503}$) in the German
sentence on the basis of having the same argument role (\textsc{safeguarder}
vs.\ \textsc{bewahrer}) in a comparable predication.

It is sometimes assumed that predicate-argument structure can be derived or
recovered from constituent structure or functional tags such as subject and
object.\footnote{See e.\,g.\ \cite{MarcusEtAl94}.} It is true that these
annotation layers provide important heuristic clues for the identification of
predicates and arguments and may eventually speed up the annotation process in
a semi-automatic way. But, as the examples above have shown,
predicate-argument structure goes beyond the assignment of phrasal categories
and grammatical functions, because the grammatical category of predicate
expressions and consequently the grammatical functions of their arguments can
vary considerably. Also, the predicate-argument structure licenses the
alignment relation by showing explicitly what it is based on.

\subsection{Binding Layer}
\label{sec:binding}
As mentioned above, we assume that the annotation model described here is used
on top of syntactically annotated data. Consequently, all elements of the
predicate-argument structure must be bound to elements of the phrasal
structure (terminal or non-terminal nodes). These bindings are stored in a
dedicated binding layer between the constituent layer and the
predicate-argument layer. 

\begin{figure*}[p]
  \begin{center}
    \epsfig{file=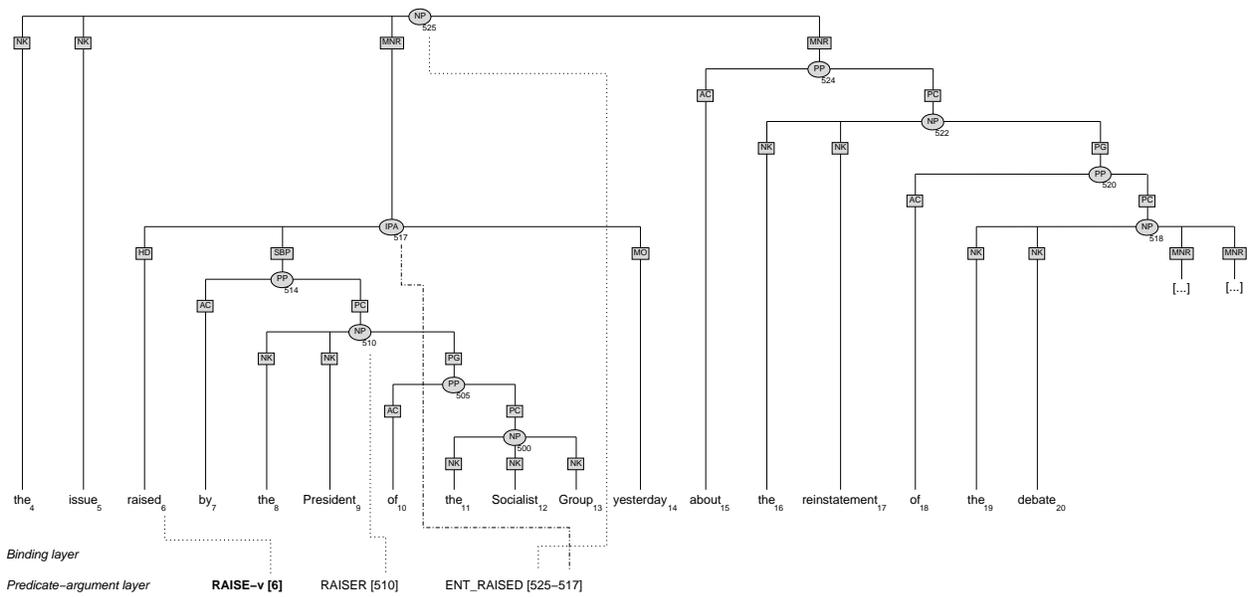,width=\textwidth}
    \caption{Complex binding of an argument: an example of a
      pruned constituent (dash-dotted line)} 
    \label{fig:binding}
  \end{center}
\end{figure*}

A problem arises when there is no direct correspondence between argument roles
and constituents. For instance, this is the case whenever a noun is
postmodified by a participle clause: in Figure~\ref{fig:binding}, the argument
role \textsc{ent\_raised} of the predicate \textsc{raise} is realised by
\textsc{np}$_{525}$, but the participle clause (\textsc{ipa}$_{517}$)
containing the predicate (\emph{raised}$_6$) needs to be excluded, because not
excluding it would lead to recursion.  Consequently, there is no simple way to
link the argument role to its realisation in the tree.

In these cases, the argument role is linked to the appropriate phrase (here:
\textsc{np}$_{525}$) and the constituent that contains the predicate
(\textsc{ipa}$_{517}$) is pruned out, which results in a discontinuous
argument realisation. Thus, in general, the binding layer allows for complex
bindings, with more than one node of the constituent structure to be included
in and sub-nodes to be explicitly excluded from a binding to a predicate or
argument.\footnote{See the database documentation \cite{Feddes04} for a more
  detailed description of this mechanism.}

When an expected argument is absent on the phrasal level due to specific
syntactic constructions, the binding of the predicate is tagged accordingly,
thus accounting for the missing argument. For example, in passive
constructions like in Table~\ref{tab:tagpred}, the predicate binding is tagged
as \dbliteral{pv}. Other common examples are imperative constructions.
Although information of this kind may possibly be derived from the constituent
structure, it is explicitly recorded in the binding layer as it has a direct
impact on the predicate-argument structure and thus might prove useful for the
automatic extraction of valency patterns.

\begin{table}[htbp]
  \centering
  \footnotesize
  \begin{tabular*}{\columnwidth}{@{\extracolsep{\fill}}l@{\quad}cccc}
    \hline
    \emph{Sentence} & wenn & korrekt   & gedolmetscht & wurde          \\
    \emph{Gloss}    & if   & correctly & interpreted  & was            \\
                    &      &           & $\uparrow$   &                \\
    \emph{Binding}  &      &           & \dbliteral{pv}  &             \\
                    &      &           & $\vert$      &                \\
    \emph{Pred/Arg} &      &           & \textsc{dolmetschen} &        \\
    \hline
  \end{tabular*}
  \caption{Example of a tagged predicate binding
    (Europarl:de-en/ep-00-01-18.al, 2532)}
  \label{tab:tagpred}
\end{table}

Note that the passive tag can also be exploited in order to query for sentence
pairs like \ref{ex:passive-e} and \ref{ex:passive-g} (in
Section~\ref{sec:voice}), where an active sentence is translated with a
passive: it is straightforward to find those instances of aligned predicates
where only one binding carries the passive tag.

\subsection{Alignment Layer}
\label{sec:align}

On the alignment layer, the elements of a pair of predicate-argument
structures are aligned with each other. Arguments are aligned on the basis of
corresponding roles within the predications.  Comparable to the tags used in
the binding layer that account for specific constructions (see
Section~\ref{sec:binding}), the alignments may also be tagged with further
information. These tags are used to classify types of non-literalness like
those discussed in Sections~\ref{sec:negation} and
\ref{sec:information}.\footnote{The deviant translations described in
  Sections~\ref{sec:nominalisations} and \ref{sec:voice} are already
  represented via predicate class (see Section \ref{sec:pa}) and on the
  binding layer (see Section \ref{sec:binding}), respectively.}

Sentences \ref{ex:inapplicable} and \ref{ex:anwendbar} are an example for a
tagged alignment.  As Section~\ref{sec:negation} has shown, negation may be
incorporated in a predicate in L1, but not in L2. Since our predicate-argument
structure does not include syntactic negation, this results in the alignment
of a predicate in L1 with its logical opposite in L2. To account for this
fact, predicate alignments of this kind are tagged as absolute opposites
(\dbliteral{abs-opp}).

\begin{figure*}[p]
  \begin{center}
    \epsfig{file=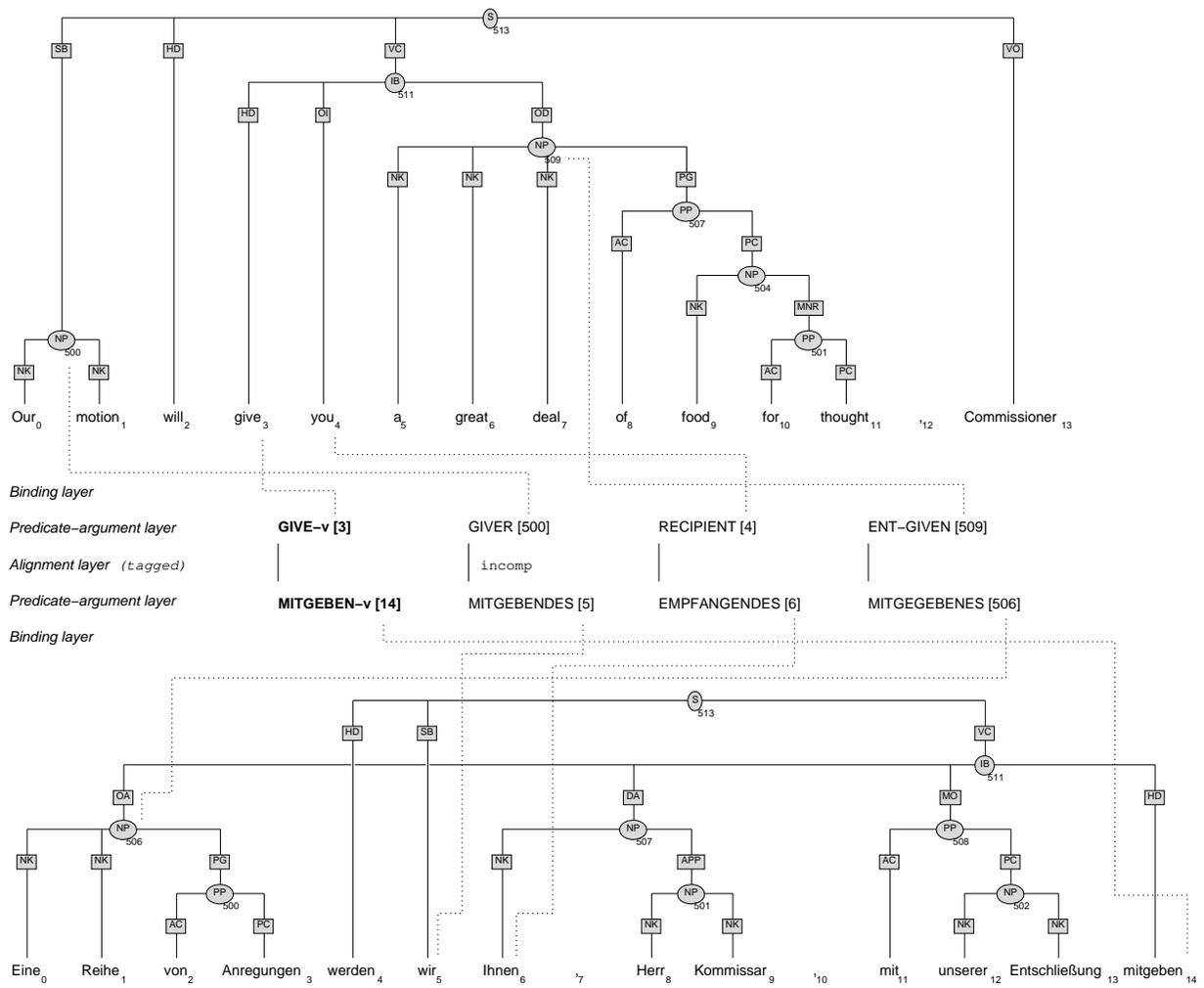,height=.55\textheight}
    \caption{Different information structure: an example of a tagged alignment
      (\dbliteral{incomp})}
    \label{fig:incompatible}
  \end{center}
\end{figure*}

Similarly, alignment tagging is applied when predications are in some way
incompatible, as is the case with sentences \ref{ex:motion} and
\ref{ex:mitgeben} in Section~\ref{sec:information}. As can be seen in the
aligned annotation (Figure~\ref{fig:incompatible}), the different information
structure of these sentences has caused the two corresponding argument roles
of \textsc{giver} and \textsc{mitgeber} to be realised by two incompatible
expressions representing different referents (\textsc{np}$_{500}$ vs.
\emph{wir}$_5$). In this case, the alignment between the incompatible
arguments is tagged \dbliteral{incomp}.

If there is no corresponding predicate-argument structure in the other
language (as e.\,g.\ the adjectival predicate in
sentence~\ref{ex:harmonisierung}) or if an argument within a structure does
not have a counterpart in the other language, there will be no alignment.

Table~\ref{tab:layers} gives an overview of the annotation layers as described
in this section.

\begin{table}[htbp]
  \centering
  \footnotesize
  \begin{tabular*}{\columnwidth}{@{}l@{\extracolsep{\fill}}l}
    \hline
    \emph{Layer} & \emph{Function}\\
    \hline
    Phrasal & constituent structure of language A\\
    Binding & binding $\downarrow$ predicates/arguments to $\uparrow$ nodes\\
    \textsc{pa} & predicate-argument structures\\
    \hline
    Alignment & aligning $\updownarrow$ predicates and arguments \\
    \hline
    \textsc{pa}& predicate-argument structures\\
    Binding & binding $\uparrow$ predicates/arguments to $\downarrow$ nodes\\
    Phrasal & constituent structure of language B\\
    \hline
  \end{tabular*}
  \caption{The layers of the predicate-argument annotation}
  \label{tab:layers}
\end{table}

All elements of the alignment structure are supposed to mark explicitly the
way they contribute to or distort the resulting translational equivalence of a
sentence pair.\footnote{Cf.\ the ``translation network'' described in
  \cite{Santos00} for a much more complex approach to describing translation
  in a formal way; this model, however, goes well beyond what we think is
  feasible when annotating large amounts of data.}  First and foremost, if two
elements are aligned to each other, this alignment is licensed by their having
comparable roles in the predicate-argument structures.  This is the default
case. If, however, a particular alignment relation, either of predicates or of
arguments, is deviant in some way, this deviance is explicitly marked and
classified on the alignment layer.

\section{Application and Outlook}
\label{sec:outlook}

The alignment model we have described is currently being used in a project to
build a treebank of aligned parallel texts in English and German with the
following linguistic levels: \textsc{pos} tags, constituent structure and
functional relations, plus the predicate-argument structure and the alignment
layer to ``fuse'' the two -- hence our working title for the treebank,
\fuse{}, which additionally stands for \emph{fu}nctional \emph{se}mantic
annotation \cite{CyrusEtAl03p,CyrusEtAl04}.

Our data source, the Europarl corpus \cite{Koehn02}, contains sentence-aligned
proceedings of the European parliament in eleven languages and thus offers
ample opportunity for extending the treebank at a later stage.\footnote{There
  are a few drawbacks to Europarl, such as its limited register and the fact
  that it is not easily discernible which language is the source language.
  However, we believe that at this stage the easy accessibility, the amount of
  preprocessing and particularly the lack of copyright restrictions make up
  for these disadvantages.} For syntactic and functional annotation we
basically adapt the \textsc{tiger} annotation scheme \cite{AlbertEtAl03},
making adjustments where we deem appropriate and changes which become
necessary when adapting to English an annotation scheme which was originally
developed for German.

We use \textsc{Annotate} for the semi-automatic assignment of \textsc{pos}
tags, hierarchical structure, phrasal and functional tags
\cite{Brants99,Plaehn98b}. \textsc{Annotate} stores all annotations in a
relational database.\footnote{For details about the \textsc{Annotate} database
  structure see \cite{Plaehn98}.} To stay consistent with this approach we
have developed an extension to the \textsc{Annotate} database structure to
model the predicate-argument layer and the binding layer.

Due to the monolingual nature of the \textsc{Annotate} database structure, the
alignment layer (Section~\ref{sec:align}) cannot be incorporated into it.
Hence, additional types of databases are needed. For each language pair
(currently English and German), an alignment database is defined which
represents the alignment layer, thus fusing two extended \textsc{Annotate}
databases.  Additionally, an administrative database is needed to define sets
of two \textsc{Annotate} databases and one alignment database. The final
parallel treebank will be represented by the union of these sets
\cite{Feddes04}.

While annotators use \textsc{Annotate} to enter phrasal and functional
structures comfortably, the predicate-argument structures and alignments are
currently entered into a structured text file which is then imported into the
database. A graphical annotation tool for these layers is under development.
It will make binding the predicate-argument structure to the constituent
structure easier for the annotators and suggest argument roles based on
previous decisions.

Possiblities of semi-automatic methods to speed up the annotation and thus
reduce the costs of building the treebank are currently being
investigated.\footnote{One track we follow is to investigate if it is feasible
  to have the annotators mark predicate-argument structures on raw texts and
  have the phrasal and functional layers added in a later stage, possibly
  supported by methods which derive these layers partially from the
  predicate-argument structures. This is, however, still very tentative.}
Still, quite a bit of manual work will remain. We believe, however, that the
effort that goes into such a gold-standard parallel treebank is very much
worthwhile since the treebank will eventually prove useful for a number of
fields and can be exploited for numerous applications.  To name but a few,
translation studies and contrastive analyses will profit particularly from the
explicit annotation of translational differences. \textsc{nlp} applications
such as Machine Translation could, e.\,g., exploit the constituent structures
of two languages which are mapped via the predicate-argument-structure. Also,
from the disambiguated predicates and their argument structures, a
multilingual valency dictionary could be derived.


\bibliographystyle{acl}
\bibliography{./paper_bib}


\end{document}